\documentclass{article}
 \usepackage[final]{neurips_2025}

\newcommand{\cmt}[1]{\textcolor{red}{\textbf{(AS: #1)}}}

\usepackage[utf8]{inputenc} 
\usepackage[T1]{fontenc}    
\usepackage{hyperref}       
\usepackage{url}            
\usepackage{booktabs}       
\usepackage{amsfonts}       
\usepackage{nicefrac}       
\usepackage{microtype}      

\usepackage{amssymb}
\usepackage{amsmath}
\usepackage{amsthm}
\usepackage{cleveref}
\usepackage{autonum}
\usepackage{MnSymbol} 

\usepackage{caption}
\usepackage{url}
\usepackage{lettrine}
\usepackage{framed}
\usepackage{color}
\usepackage{mathtools}

\usepackage{algorithm,algpseudocode}

\algnewcommand{\algorithmicforeach}{\textbf{for each}}
\algdef{SE}[FOR]{ForEach}{EndForEach}[1]
  {\algorithmicforeach\ #1\ \algorithmicdo}
  {\algorithmicend\ \algorithmicforeach}

\usepackage{overpic}
\usepackage{dsfont}
\usepackage{booktabs} 
\usepackage{bm}
\usepackage{bbm}
\usepackage{ifthen}
\usepackage{graphicx}
\usepackage{subcaption}
\usepackage{dsfont}
\usepackage{framed}
\usepackage{overpic}
\usepackage[svgnames]{xcolor}
\newtheorem{theorem}{Theorem}
\usepackage{apxproof}

\newtheorem{definition}{Definition}
\newtheorem{lemma}{Lemma}
\newtheorem{assumption}{Assumption}

\hypersetup{
	colorlinks=true,
	linkcolor=blue,
	urlcolor=red,
	citecolor=DarkBlue,
	linkbordercolor={0 0 1}
}

\title{Beyond $\tilde{O}(\sqrt{T})$ Constraint Violation for Online Convex Optimization with Adversarial Constraints}
\author{%
  Abhishek Sinha \quad  Rahul Vaze \\
 School of Technology and Computer Science \\
  Tata Institute of Fundamental Research \\
  Mumbai 400005, India \\
  \texttt{abhishek.sinha@tifr.res.in}, 
  \texttt{rahul.vaze@gmail.com}
}

\begin{document}

\maketitle
\begin{abstract}
We study Online Convex Optimization with adversarial constraints (COCO). At each round a learner selects an action from a convex decision set and then an adversary reveals a convex cost and a convex constraint function.
The goal of the learner is to select a sequence of actions to minimize both regret and the cumulative constraint violation (CCV) over a horizon of length $T$. The best-known policy for this problem achieves $O(\sqrt{T})$ regret and $\tilde{O}(\sqrt{T})$ CCV.  
In this paper, we improve this by trading off regret to achieve substantially smaller CCV.
This trade-off is especially important in safety-critical applications, where satisfying the safety constraints is non-negotiable. Specifically, for any bounded convex cost and constraint functions, we propose an online policy that achieves $\tilde{O}(\sqrt{dT}+ T^\beta)$ regret and $\tilde{O}(dT^{1-\beta})$ CCV, where $d$ is the dimension of the decision set and $\beta \in [0,1]$ is a tunable parameter. We begin with a special case, called the \textsc{Constrained Expert} problem, where the decision set is a probability simplex and the cost and constraint functions are linear. Leveraging a new adaptive small-loss regret bound, we propose a computationally efficient policy for the \textsc{Constrained Expert} problem, that attains $O(\sqrt{T\ln N}+T^{\beta})$ regret and $\tilde{O}(T^{1-\beta} \ln N)$ CCV for $N$ number of experts. The original problem is then reduced to the \textsc{Constrained Expert} problem via a covering argument. Finally, 
with an additional $M$-smoothness assumption, we propose a computationally efficient first-order policy attaining $O(\sqrt{MT}+T^{\beta})$ regret and $\tilde{O}(MT^{1-\beta})$ CCV.
\end{abstract}
\section{Introduction}

Online Convex Optimization (OCO) is a standard framework for sequential decision-making under adversarial uncertainty \citep{hazan2022introduction}. At each round $1 \leq t \leq T$, a learner selects an action $x_t$ from a convex decision set $\mathcal{X}$ with finite diameter $D$. The environment then reveals a convex, Lipschitz continuous  cost function $f_t: \mathcal{X} \mapsto \mathbb{R}$, and the learner incurs a cost of $f_t(x_t)$. The objective is to minimize the regret relative to the best fixed action in hindsight. For any comparator action $x^\star \in \mathcal{X},$ the regret is defined as: \begin{eqnarray} \label{reg-def1} 
\textrm{Regret}_T(x^\star) = \sum_{t=1}^T f_t(x_t) - \sum_{t=1}^T f_t(x^\star). 
\end{eqnarray} The worst-case regret is defined to be $\sup_{x^\star \in \mathcal{X}} \textrm{Regret}_T(x^\star)$. It is well-known that simple algorithms such as Online Mirror Descent (OMD) attain an $O(\sqrt{T})$ worst-case regret, which is also minimax optimal \citep{hazan2022introduction}. 


Online Convex Optimization with adversarial constraints (COCO) generalizes the standard OCO framework and underpins a variety of emerging applications, including AI safety \citep{amodei2016concrete, pmlr-v70-sun17a}, fair allocation \citep{sinha2023banditq}, online ad markets with budget constraints \citep{georgios-cautious}, and multi-task learning \citep{ruder2017overview, dekel2006online}. In COCO, on each round $1\leq t \leq T,$ the learner selects an action $x_t$ from a convex decision set $\mathcal{X}$ with diameter $D$. 
 The adversary then reveals two convex functions - a cost function $f_t: \mathcal{X} \mapsto \mathbb{R}$ and a constraint function $g_t: \mathcal{X} \mapsto \mathbb{R}.$ For simplicity, we make the following mild assumption:
\begin{assumption}[Bounded Cost and Constraints] \label{bdd-assump}
	The cost and constraint functions are bounded within their domain. In particular, via appropriate translation and scaling, we assume that
 $0\leq f_t, g_t \leq 1, \forall t.$
\end{assumption}
  See Section \ref{bdd} in the Appendix for a brief discussion on Assumption \ref{bdd-assump}. The constraint function $g_t$ corresponds to a constraint of the form $h_t(x)\leq 0$ where we define $g_t(x) \equiv \max(0, h_t(x)).$ Thus, $g_t(x_t)$ quantifies the penalty incurred by the learner for violating the hard constraint of $h_t(x) \leq 0$ at round $t$. If an action $x^\star$ satisfies $g_t(x^\star) = 0, \forall t$, then it is feasible throughout the horizon. To measure long-term constraint violation of a policy, we define Cumulative Constraint Violation (CCV) as: 
\begin{eqnarray}  \label{ccv-def}
\textrm{CCV}_T = \sum_{t=1}^T g_t(x_t). 
\end{eqnarray}


Since the constraint on each round is revealed \emph{after} the learner selects action - and may be chosen adversarially - it is generally impossible for an online policy to satisfy the constraints on every round. Therefore, to ensure the problem is well-posed, one must impose some restrictions on the constraint functions \citep{mannor2009online}. In the COCO literature, the following feasibility assumption is universally made \citep{sinha2024optimal, guo2022online, neely2017online, yuan2018online, yi2021regret}. 

We define the feasible set $\mathcal{X}^\star$ to be the subset of actions that satisfy the constraints across \emph{all rounds}:
\begin{eqnarray} \label{feas-set}
	\mathcal{X}^\star = \{x \in \mathcal{X}: g_t(x) = 0, ~\forall t \geq 1\}.
\end{eqnarray}  
\begin{assumption}[Feasibility] \label{feasibility}
	The feasible set is non-empty, \emph{i.e.,} $\mathcal{X}^\star \neq \emptyset.$ 
	\end{assumption}
	The feasibility assumption is not essential for our results. In particular, our algorithmic and analytical techniques naturally extend to the more general setting where the feasible actions are allowed to violate the constraints up to a prescribed budget of $B_T\geq 0$ \citep{sarkar2025online}. Please refer to Appendix \ref{relaxed_feasibility} and Theorem \ref{cvx-smooth-lt-thm} for this generalization.  
	
	In the COCO problem, the regret of any policy is computed relative to the best feasible action in hindsight, \emph{i.e.,}
\begin{eqnarray} \label{worst-reg-def}
	\textrm{Regret}_T = \sup_{x^\star \in \mathcal{X}^\star} \textrm{Regret}_T(x^\star).
\end{eqnarray}

\paragraph{Goal:}The standard objective in the COCO problem is to design an online policy that achieves both small regret and small cumulative constraint violation (CCV). In this work, our primary focus is on designing a flexible framework that minimizes the CCV to the extent possible, while ensuring that the regret remains sublinear in the horizon length $T$. This trade-off is particularly important in safety-critical applications, such as autonomous driving, where reducing constraint violation (\emph{e.g.,} safety breaches) takes precedence over minimizing regret (\emph{e.g.,} fuel or battery optimization, commute time reduction).
%
\subsubsection*{Background and Our Contribution}
Recently, \citet{sinha2024optimal} proposed a computationally efficient first-order policy for the COCO problem, which achieves $O(\sqrt{{T}})$ regret and $\tilde{O}(\sqrt{T})$ CCV. They also established the tightness of their results in the high-dimensional regime, where the dimension $d$ of the decision set $\mathcal{X}$ is at least $T.$ It is well-known that even in the standard OCO problem, where $g_t =0, \forall t$, the regret is lower bounded by $\Omega(\sqrt{T}),$ even for $d=1$ \citep[Theorem 3.2]{hazan2022introduction}. Thus the unconditional $O(\sqrt{T})$ regret guarantee for COCO cannot be improved. However, the question of whether one can achieve a CCV substantially smaller than $\tilde{O}(\sqrt{T})$ under additional natural assumptions was left open.

 Our main contribution in this paper is to affirmatively answer this question. In particular, we show that in the fixed-dimensional setting where $d \ll T,$ it is possible to achieve significantly smaller cumulative constraint violation (CCV) while appropriately trading off the regret.
 Furthermore, when the cost and constraint functions are smooth, we show that a computationally efficient online gradient descent-based policy can achieve improved guarantees. A summary of our results is provided in Table~\ref{gen-oco-review-table}. From the table, we also note that with an appropriate choice of the parameters ($\beta=1$), our proposed policy achieves $O(\ln T)$ CCV in the special case of \textsc{Online Constraint Satisfaction} (OCS) problem when all cost functions are zero and the only goal is to satisfy the constraints \citep{sinha2024optimal}.

A key technical ingredient in our analysis is the use of \emph{small-loss} regret bounds, also known as $L^\star$ bounds in the online learning literature \citep{cesa2006prediction, orabona2019modern}. These bounds yield guarantees that improve upon the minimax optimal $O(\sqrt{T})$ rate for standard regret minimization problem when the fixed comparator incurs a small cumulative loss. In Section ~\ref{prelims}, we extend these classical results by proposing a new adaptive policy for the \textsc{Expert} problem that achieves a small-loss regret bound in the general setting where the per-round loss vectors can be potentially \emph{unbounded}. In Section \ref{simplex}, we consider a special case of COCO, called the \textsc{Constrained Expert} problem, 
where the decision set is an $N$-dimensional simplex and the cost and constraint functions are linear. In Section \ref{gen_set}, we reduce the general COCO problem 
to the \textsc{Constrained Expert} problem via a covering argument. In Section \ref{cvx_case}, we give a computationally efficient first-order policy with improved bounds for smooth and convex functions. 
Due to space constraints, experimental results have been deferred to Section \ref{expt} in the Appendix.

 \begin{table*}[!ht]
  \title{Summary of the results for the constrained OCO problem}
  \begin{tabular}{llllll}
    \toprule
   \small { Reference}  & \small {Regret} & \small {CCV} & \small {Complexity}& \small {Assumptions} \\
    \midrule
    \small {\citet{jenatton2016adaptive}}  & \small {$O(T^{\max(\beta, 1-\beta)})$} & \small {$O(T^{1-\beta/2})$} & \small {Projection} & \small{\textsf{FC}} \\
    \small {\citet{neely2017online}}  & \small {$O(\sqrt{T})$} & \small {$O(\sqrt{T}/\eta)$} & \small {Conv-OPT} & \small {Slater condition} \\
  \small {\citet{yuan2018online}} & \small {$O(T^{\max(\beta, 1-\beta)})$} & \small {$O(T^{1-\beta/2})$ }  & \small {Projection} & \small{\textsf{FC}} \\
      \small {\citet{yu2020low}}  & \small {$O(\sqrt{T})$} & \small {$O(1/\eta)$} & \small {Conv-OPT} & \small {Slater, \textsf{FC}} \\
  \small {\citet{yi2021regret}} & \small {$O(T^{\max(\beta, 1-\beta)})$} & \small {$O(T^{(1-\beta)/2})$} & \small {Conv-OPT} & \small{\textsf{FC}} \\ 
    \small {\citet{guo2022online}}  & \small {$O(\sqrt{T})$} & \small {$O(T^{\nicefrac{3}{4}})$} & \small {Conv-OPT} & - \\
  \small {\citet{yi2023distributed}}  & \small {$O(T^{\max(\beta, 1-\beta)})$} & \small {$O(T^{1-\beta/2})$} & \small {Conv-OPT} & - \\
     \small {\citet{sinha2024optimal}} & \small {$O(\sqrt{T})$} & \small {$\tilde{O}(\sqrt{T})$} & \small {Projection} & -\\
      \small {\citet{vaze2025sqrt}} & \small {$O(\sqrt{T})$} & \small {Instance dependent} & \small {Projection} & -\\
  \small {\textbf{This paper}} & \small {$O(\sqrt{T\ln N}+T^{\beta})$} & \small {$\tilde{O}(T^{1-\beta}\ln N)$} & \small {$O(N)$} & \small{\textsc{Constr. Expert}}\\
    \small {\textbf{This paper}} & \small {$\tilde{O}(\sqrt{dT}+ T^\beta)$} & \small {$\tilde{O}(dT^{1-\beta})$} & \small {$O(T^d)$} & \small{$d$-dim. decision set}\\
          \small {\textbf{This paper}} & \small {$O(\sqrt{MT}+ T^\beta)$} & \small {$\tilde{O}(MT^{1-\beta})$} & \small {$\textrm{Projection}$} & \small{
          Smooth}\\
       \bottomrule
  \end{tabular}
  \vspace{5pt}
  \caption{\small{Summary of the key results on the COCO problem. In the above table, $\beta \in[0,1]$ is a tunable parameter, $\eta >0$ denotes the Slater's constant, $N$ denotes the number of experts, $d$ denotes the dimension of the decision set $\mathcal{X}$, $M$ denotes the smoothness constant, and $\tilde{O}(\cdot)$ hides polylogarithmic factors in $T$. \textsc{Constr. Expert} refers to the \textsc{Constrained Expert} problem described in Section \ref{simplex}, Conv-OPT refers to solving a constrained convex optimization problem on each round, Projection refers to the Euclidean projection operation on the decision set $\mathcal{X}$, and \textsf{FC} refers to the Fixed Constraints setting.} }
    \label{gen-oco-review-table}
\end{table*}

\paragraph{Intuition for the results:} We now give some intuition for why the Cumulative Constraint Violation (CCV) can be expected to be made smaller than the current-best bound of $\tilde{O}(\sqrt{T})$ by crucially utilizing small-loss regret bounds. Consider the Online Constraint Satisfaction (OCS) problem, introduced by \citet{sinha2024optimal}, where all cost functions are identically equal to zero and the goal is to minimize the CCV only. Let the constraint functions (\emph{i.e.,} $\{g_t\}_{t\geq 1}$'s), be non-negative, $M$-smooth, and convex. 

For solving this problem, we run the Online Gradient Descent policy on the sequence of constraint functions with an adaptive step size schedule. Specifically, we choose the next action as 
$x_{t+1} = \textsc{Proj}_{\mathcal{X}}(x_t - \eta_t \nabla_t),$
where $\nabla_t \equiv \nabla g_t(x_t)$ and the step sizes are adaptively chosen as $\eta_t = D/\sqrt{2\sum_{\tau=1}^t || \nabla_\tau||_2^2}, t\geq 1.$ The following small-loss regret bound achieved by this policy for non-negative smooth functions is well-known \citep[Theorem 4.25]{orabona2019modern} (please see the statement of Theorem \ref{ogd-smooth} in Section \ref{cvx_case} for a quick reference):
\begin{eqnarray} \label{smooth-reg-1}
	\sum_{t=1}^T g_t(x_t) - \sum_{t=1}^T g_t(u) \leq 4D^2M + 4D \sqrt{M \sum_{t=1}^T g_t(u)}, ~\forall u \in \mathcal{X}. 
\end{eqnarray}
Now, if we choose the comparator $u$ to be a feasible action by setting $u=x^\star, x^\star \in \mathcal{X}^\star$, we have $g_t(x^\star)=0, \forall t. $ Thus the regret bound \eqref{smooth-reg-1} implies  $\textrm{CCV}_T = \sum_{t=1}^T g_t(x_t) \leq 4D^2M$ -  which is a \emph{constant} independent of $T.$ This result is surprising as the $O(\sqrt{T})$ lower bound for regret holds even for linear functions \citep{hazan2022introduction}.

 In this paper, we generalize this observation by exploring the trade-off between regret and CCV while taking into account both the cost and constraint functions. Our main result roughly says that any $(\textrm{Regret}_T, \textrm{CCV}_T)$ pair with $\textrm{Regret}_T \geq \tilde{\Theta}(\sqrt{T})$ and $\textrm{Regret}_T \times \textrm{CCV}_T = \tilde{O}(T)$ is achievable.
 


\subsection{Prior work}
Online convex optimization with constraints has been extensively studied under various modeling assumptions. Table~\ref{gen-oco-review-table} provides a summary of key results in the literature.

In the fixed-constraint setting, where \( g_t = g \) for all \( t \), \citet{yi2021regret} proposed a policy that achieves \( O(T^{\max(\beta, 1 - \beta)}) \) regret and \( O(T^{(1 - \beta)/2}) \) CCV. Under the stronger assumption of \emph{Slater’s condition}, which requires the existence of a uniformly strictly feasible action \( x^\star \in \mathcal{X} \) such that \( g_t(x^\star) \leq -\eta \) for some constant \( \eta > 0 \) and all \( t \), \citet{yu2020low} showed that the CCV can be reduced to \( O(1/\eta) \) while maintaining \( O(\sqrt{T}) \) regret.

The problem becomes significantly more challenging in the presence of time-varying adversarial constraints. Under Slater’s condition, \citet{neely2017online} developed an algorithm achieving \( O(\sqrt{T}) \) regret and \( O(\sqrt{T}/\eta) \) CCV. However, since  this condition is quite strong, difficult to verify in practice, and leads to vacuous CCV bounds as \( \eta \to 0^+ \), recent works have focused on avoiding this assumption.

\citet{guo2022online} proposed an algorithm that needs to solve a separate offline convex optimization problem in each round, achieving \( O(\sqrt{T}) \) regret and \( O(T^{3/4}) \) CCV. Subsequently, \citet{sinha2024optimal} introduced a simpler gradient-based policy that improves the CCV bound to \( \tilde{O}(\sqrt{T}) \) while still maintaining \( O(\sqrt{T}) \) regret. See also \citet{lekeufack2024optimistic, lu2025order, supantha2025universal, sarkar2025online} for various extensions of their result. More recently, \citet{vaze2025sqrt} proposed an algorithm that achieves \( O(\sqrt{T}) \) regret 
and constant CCV for some special classes of the constraint sets.

The algorithms proposed in this paper make use of Lipschitz-adaptive small-loss regret bounds. Small-loss regret bounds for the standard regret minimization problem with \emph{bounded} Lipschitz constants, where the regret scales with the cumulative loss of the benchmark (instead of the time horizon $T$), have been well studied \citep{cesa1997use, auer2002adaptive, hazan2010extracting}. \citet{mhammedi2019lipschitz} propose Lipschitz-adaptive algorithms in the \textsc{Expert} setting, which rely on multiple restart phases and incur additional computational overhead. Similarly, \citet{cesa2007improved} uses doubling trick to estimate the unknown parameters. Restarting learning algorithms during their course of execution is practically wasteful as it discards all past data prior to the restarts. To the best of our knowledge, continuously adaptive, scale-free variants of the small-loss regret bounds in the \textsc{Expert} setting have not been investigated before in the literature. In the following, we review the \textsc{Expert} problem and derive an adaptive small-loss regret bound. 

\section{Preliminaries: Adaptive Small-Loss Regret Bound for the \textsc{Expert} Problem} \label{prelims}
The \emph{Prediction with Expert Advice} problem, also known as the \textsc{Expert} problem in the literature, refers to a repeated game where, in each round $t\geq 1,$ the learner chooses a probability distribution $p_t$ over a set of $N$ experts (experts may be identified with the set of actions). After that, the adversary chooses a bounded loss vector $l_t \in [0,1]^N,$ where $l_t(i)$ denotes the loss for the $i$\textsuperscript{th} expert, $i \in [N]$. Consequently, the learner incurs an expected loss of $f_t(p_t)=\langle l_t, p_t\rangle$ in round $t.$ The full loss vector $l_t$ is revealed to the learner at the end of round $t.$ The learner's objective is to choose a sequence of distributions $\{p_t\}_{t \geq 1}$ to minimize its regret over $T$ rounds. The \textsc{Expert} problem is the full-information counterpart of the Multi-armed Bandit (MAB) problem \citep{bubeck2012regret}.

The Exponential Weights algorithm, also known as Hedge, is a well-known solution to the \textsc{Expert} problem. The Hedge algorithm selects the $i$\textsuperscript{th} expert on round $t$ with probability 
\begin{eqnarray} \label{hedge-prob}
	p_t(i) = \exp\big(-\eta \sum_{\tau=1}^{t-1}l_\tau(i)\big)/Z, ~i\in [N],
\end{eqnarray}
 where $\eta>0$ is a  suitably chosen learning rate and $Z$ is the normalizing constant. Hedge is known to achieve the following \emph{small-loss} regret bound  
\citep[Corollary 2.4]{cesa2006prediction}:
 \begin{eqnarray} \label{small-loss}
 	\textrm{Regret}_T = \tilde{L}_T- L_T(i^\star) \leq \sqrt{2L_T(i^\star) \ln N} + \ln N,
 \end{eqnarray} 
where $\tilde{L}_T = \sum_{\tau=1}^T \langle l_\tau, p_t\rangle $ is the learner's cumulative loss,  $L_T(i^\star) = \sum_{\tau=1}^T l_\tau(i^\star)$ is the cumulative loss of a comparator arm $i^\star \in [N]$. See also the Squint algorithm proposed by \citet{koolen2015second}.  
The small-loss bound \eqref{small-loss} offers a significant improvement over the standard $O(\sqrt{T})$ regret bound when the comparator's loss is small, \emph{i.e.,} $L_T(i^\star) =o(T)$. While the bound \eqref{small-loss} is well-known, it assumes that that all loss values are bounded \citep[Section 2.4]{cesa2006prediction}. 
However, for reasons that will become clear in the sequel, we cannot use \eqref{small-loss} in our algorithm, as we will need to control the regret when the losses, defined appropriately by the algorithm, are unbounded.

To address this technical challenge, Theorem \ref{gen-small-loss} presents a small-loss regret bound for the \textsc{Expert} problem that generalizes \eqref{small-loss} by accommodating potentially unbounded losses. This regret bound is achieved by an adaptive variant of the Hedge policy that employs a self-confident variable learning rate. The full pseudocode for this policy is provided in Algorithm \ref{adapt_hedge} in Appendix \ref{adapt-hedge-pseudocode}. 

\begin{theorem} \label{gen-small-loss}
 Consider the \textsc{Expert} problem with $N$ experts. Let the vector $l_t$ denote the losses of the experts at round $t \geq 1,$ where $l_t(i)\geq 0, \forall i,t.$ The losses need not be uniformly bounded above for all rounds. Let $G_t$ be an upper bound to $||l_t||_\infty$ satisfying the following conditions\footnote{Naturally, the $\{G_t\}_{t\geq 1}$ sequence is causal \emph{i.e.,} $G_t$ can be computed only after observing the loss vector $l_t$ on round $t.$}: 
 \begin{enumerate}
 	\item The sequence $\{G_t\}_{t\geq 1}$ is monotonically non-decreasing, \emph{i.e.,} $G_{t} \geq G_{t-1}, \forall t\geq 1, G_0=1.$ 
 	\item The growth of $G_t$ in consecutive rounds is bounded: $\max_{1\leq t\leq T} \frac{G_t}{G_{t-1}} \leq \gamma$ for some known constant $\gamma \geq 1.$ 
 	\end{enumerate}
 Then the following adaptive Hedge algorithm, which selects the $i$\textsuperscript{th} expert with probability 
$p_t(i) \propto \exp(-\eta_t L_{t-1}(i)), \forall i,$ with an adaptive learning rate $\eta_t = \frac{1}{\sqrt{G_{t-1}}}\sqrt{\frac{\ln N}{\tilde{L}_{t-1}+ \gamma G_{t-1}}},$
achieves the following regret bound: 
\begin{eqnarray} \label{adapt-hedge-reg-bd}
  \tilde{L}_T- L_T(i^\star) \leq 2\gamma \sqrt{L_T(i^\star) G_T \ln N } +  7 \gamma^2 G_T \ln N. 	
\end{eqnarray} 
 In the above, $\tilde{L}_{t} = \sum_{\tau=1}^{t} \langle p_\tau, l_\tau \rangle$ denotes the algorithm's cumulative loss up to round $t$ and $L_T(i^\star) = \sum_{\tau=1}^T l_\tau(i^\star)$ is the cumulative loss of any comparator expert $i^\star \in [N]$ up to round $T.$
\end{theorem}
 See Appendix \ref{gen-small-loss-proof} for a  proof of Theorem \ref{gen-small-loss}.
 
\textbf{Remarks:} 1. The monotonicity assumption on the sequence $\{G_t\}_{t\geq 1}$ entails no loss of generality, since one can always define a new sequence of upper bounds as $G_t' := \max_{1\leq \tau \leq t} G_\tau, t \geq 1,$ which is monotonic by construction.

2. The proof of Theorem~\ref{gen-small-loss} is non-trivial because the value of \( G_T \) is unknown in advance. As a result, we cannot simply rescale all losses by \( G_T \) and directly apply the small-loss bound in \eqref{small-loss} to the normalized losses. Instead, we carefully design an adaptive learning rate schedule that accounts for the growth of the loss scale and the cumulative loss of the best expert over time.

\section{The \textsc{Constrained Expert} problem: Simplex Decision Set, Linear Cost, and Linear Constraint Functions} \label{simplex}
In this section, we focus on an important special case of the COCO problem, called \textsc{Constrained Expert} problem, where the cost and constraint functions are linear and the decision set is the $N-1$-dimensional simplex $\Delta_N$, \emph{i.e.,}
 \begin{eqnarray*}
 	\mathcal{X} = \Delta_N= \left \{p \in \mathbb{R}^N: \sum_{i=1}^N p_i =1, p_i \geq 0, \forall i \right \}.
 \end{eqnarray*}
 In this problem, at each round $t,$ the learner first computes a distribution $p_t \in \Delta_N$ over $N$ experts, and then it samples an expert from the distribution $p_t$.
Choosing expert $i$ incurs a cost of $f_t(i)$ and a constraint violation of $g_t(i).$ Consequently, the learner incurs an expected cost of $f_t(p_t) \equiv \langle f_t, p_t \rangle$ and an expected constraint violation of $g_t(p_t) \equiv \langle g_t, p_t\rangle$ on round $t$\footnote{To simplify the notations, we use the same symbol for denoting a linear function and its associated coefficient vector.}. Both the cost vector $f_t$ and the constraint vector $g_t$ are revealed to the learner at the end of the round. The feasibility assumption (Assumption \ref{feasibility}), specialized to this setting, implies that there exists at least one expert $i^\star \in [N]$ such that $g_t(i^\star)=0, \forall t.$ The learner's objective is to generate a sequence of distributions $\{p_t\}_{t=1}^T$ that minimizes both the regret and the cumulative constraint violation (CCV).

\subsection{An Online Policy for the \textsc{Constrained Expert} Problem} \label{constr-sec}
Let $\Phi: \mathbb{R} \mapsto \mathbb{R}$ be a non-decreasing convex Lyapunov function, to be specified later, and let $i^\star$ be an uniformly feasible expert with $g_t(i^\star)=0, \forall t$. As stated earlier, the existence of $i^\star$ is guaranteed by Assumption \ref{feasibility}. Let $Q(t)$ denote the cumulative constraint violation (CCV) up to round $t,$ which evolves as follows:
 \begin{eqnarray} \label{q-ev3}
 	Q(t)= Q(t-1)+  \langle g_t, p_t \rangle.  
 \end{eqnarray}
We now use the regret decomposition framework introduced by \citet{sinha2024optimal} to design an online policy for the \textsc{Constrained} \textsc{Expert} problem.
Using the convexity of the Lyapunov function $\Phi(\cdot),$ we have for any $1\leq t \leq  T$:
\begin{eqnarray} \label{phi-q}
	\Phi(Q(t)) - \Phi(Q(t-1)) &\leq& \Phi'(Q(t))(Q(t)-Q(t-1)) \nonumber\\
	&\stackrel{(a)}{=}&  \Phi'(Q(t)) \langle g_t, p_t\rangle \nonumber\\
	&\stackrel{(b)}{=}& \Phi'(Q(t)) (\langle g_t, p_t\rangle - g_t(i^\star)), 
\end{eqnarray}
where (a) follows from Eqn.\ \eqref{q-ev3} and (b) uses the fact that $g_t(i^\star)=0, \forall t.$ Adding the term $\langle f_t, p_t \rangle - f_t(i^\star)$ to both sides of the inequality, we obtain  
\begin{eqnarray} \label{reg-decomp1}
	&&\Phi(Q(t)) - \Phi(Q(t-1)) + (\langle f_t, p_t \rangle - f_t(i^\star)) \nonumber\\
	&&\leq \langle f_t + \Phi'(Q(t)) g_t, p_t \rangle - (f_t(i^\star) + \Phi'(Q(t)) g_t(i^\star)).
\end{eqnarray}
Define the $t$\textsuperscript{th} surrogate cost function $\hat{f}_t : \Delta_N \mapsto \mathbb{R}$ to be the following linear function:
\begin{eqnarray} \label{surr-cost}
	\hat{f}_t(p) \coloneqq \langle f_t + \Phi'(Q(t)) g_t, p\rangle, ~~ t\geq 1.
\end{eqnarray}
Summing up Eqn.\ \eqref{reg-decomp1} for $1\leq t \leq T, $ we obtain the following regret decomposition inequality:
\begin{eqnarray} \label{reg-decomp-ineq}
	\Phi(Q(T))- \Phi(Q(0)) + \textrm{Regret}_T(i^\star) \leq \textrm{Regret}'_T(i^\star).
\end{eqnarray}
In Eqn.\ \eqref{reg-decomp-ineq}, $\textrm{Regret}_T(i^\star) \equiv \sum_{t=1}^T \langle f_t, p_t-e_{i^\star} \rangle$ and $\textrm{Regret}_T'(i^\star) \equiv \sum_{t=1}^T \langle \hat{f}_t, p_t-e_{i^\star} \rangle$ correspond to the regret for the original cost functions $\{f_t\}_{t=1}^T$ and the surrogate cost functions $\{\hat{f}_t\}_{t=1}^T,$ respectively, with respect to a feasible expert $i^\star \in [N].$

 Algorithm \ref{coco_alg} presents our proposed online policy for the \textsc{Constrained Expert} problem. It employs the adaptive Hedge subroutine from Theorem \ref{gen-small-loss} to select the sampling distributions $\{p_t\}_{t\geq 1}$ which minimize $\textrm{Regret}_T'(i^\star)$ appearing on the RHS of Eqn.\ \eqref{reg-decomp-ineq}. Observe that the surrogate cost function $\hat{f}_t$ involves the term $\Phi'(Q(t)),$ which may grow indefinitely with $t$ as $Q(t)$ grows. Hence, it is imperative to use the adaptive Hedge algorithm (Algorithm \ref{adapt_hedge}) which can handle unbounded loss, in contrast to the standard Hedge algorithm \eqref{hedge-prob},
  which assumes bounded loss vectors. The following theorem gives an upper bound to the regret and CCV achieved by Algorithm \ref{coco_alg}.

 \begin{algorithm}[tb]
   \caption{Algorithm for the \textsc{Constrained Expert} problem}
   \label{coco_alg}
\begin{algorithmic}[1]
   \State {\bfseries Input:} Number of experts $N$, Horizon length $T$, Cost vectors $\{f_t\}_{t=1}^T$ and  Constraint vectors $\{g_t\}_{t=1}^T,$ Tunable parameter $\beta \in [0,1].$ 
         \State {\bfseries Parameter settings:} 
      $\Phi(x)= e^{\lambda x}, \lambda = T^{-(1-\beta)}/(2c\ln N), c=10.$
  \State {\bfseries Initialization:} Set $Q(0) \gets 0,$ Algorithm's loss $\tilde{L}_0 \gets 0$, Experts' cumulative loss $L_0 \gets \bm{0}, G_0=1+\Phi'(Q(0)).$ 
   \ForEach{$t=1:T$}
   \State Compute self-confident learning rate
        \begin{eqnarray} \label{learning-rate}
 	\eta_t = \frac{1}{\sqrt{G_{t-1}}}\sqrt{\frac{\ln N}{\tilde{L}_{t-1}+ \gamma G_{t-1}}}.
 \end{eqnarray} 
    \State Play adaptive Hedge by choosing the $i$\textsuperscript{th} expert with probability
    \begin{eqnarray*}
    	    p_t(i) = \exp(-\eta_t L_{t-1}(i))/\sum_{j=1}^N \exp(-\eta_t L_{t-1}(j)), ~\forall i \in [N].
    \end{eqnarray*}
   \State Observe the vectors $f_t, g_t.$ Incur a cost of $\langle f_t, p_t \rangle $ and constraint violation of $ \langle g_t, p_t \rangle.$ 
     \State Update CCV and $G_t$: 
     \begin{eqnarray*}
     	      Q(t)=Q(t-1)+\langle g_t, p_t \rangle, ~~
     	      G_t = 1+ \Phi'(Q(t)).
     \end{eqnarray*}
   \State Compute the surrogate cost vector: 
   	   $\hat{f}_t =  f_t + \Phi'(Q(t)) g_t.$ (coordinate-wise vector addition)
   \State Update the cumulative loss of the algorithm and the cumulative loss of each expert 
   \begin{eqnarray*}
   	\tilde{L}_t = \tilde{L}_{t-1} + \langle \hat{f}_t, p_t \rangle,~~ 
   	L_t(i)= L_{t-1}(i) + \hat{f}_t(i), ~~i \in [N].
   \end{eqnarray*} 
   \EndForEach
\end{algorithmic}
\end{algorithm}

 \begin{theorem} \label{main-th}
 	Consider the Lyapunov function $\Phi(x)=e^{\lambda x},$ where $ \lambda = T^{-(1-\beta)}/(2c\ln N), c=10,$ and $\beta \in [0,1]$ is a tunable parameter. Then, under Assumptions \ref{bdd-assump} and \ref{feasibility}, Algorithm \ref{coco_alg} achieves the following guarantees for the \textsc{Constrained Expert} problem for any feasible expert $i^\star \in [N]$:
 	\begin{eqnarray*}
 	\textrm{Regret}_T(i^\star) = O(\sqrt{T\ln N}+ T^\beta +\ln N), ~ \textrm{CCV}_T =O(T^{1-\beta}\ln N \ln T). 
 	\end{eqnarray*}
 	In particular, if $f_t=0, \forall t,$ upon setting $\beta=1,$ we obtain $\textrm{CCV}_T= O(\ln N \ln T).$
 \end{theorem}
 Experimental results, presented in Section \ref{expt} of the Appendix, qualitatively support the expected variations of Regret and CCV as a function of the parameter $\beta$.
\subsection{Proof of Theorem \ref{main-th}}
 Define the sequence $G_t \equiv 1+\Phi'(Q(t)), t \geq 0,$ where the Lyapunov function $\Phi(\cdot)$ is chosen as specified in the statement of Theorem \ref{main-th}. From the definition of the surrogate costs (Eqn.\ \eqref{surr-cost}), it follows that $G_t$ is an upper bound to the maximum component of the surrogate cost vector: \[||\hat{f}_t||_\infty \leq ||f_t||_\infty + \Phi'(Q(t)) ||g_t||_\infty \leq 1+ \Phi'(Q(t))= G_t, \forall t.\] Since $Q(t)$ is non-decreasing in $t$ and the function $\Phi(\cdot)$ is convex, it follows that the sequence $\{G_t\}_{t \geq 1}$ is also non-decreasing. Furthermore, from Lemma \ref{ratio-bd} in the Appendix, we have that  
 $ \max_{1 \leq t \leq T} \frac{G_t}{G_{t-1}}\leq 1.08$. Thus all conditions in Theorem \ref{gen-small-loss} are fulfilled with $\gamma= 1.08,$ and we can use the small-loss regret bound \eqref{adapt-hedge-reg-bd} of the adaptive Hedge algorithm for the surrogate cost sequence $\{\hat{f}_t\}_{t=1}^T.$
Let $i^\star \in [N]$ be a uniformly feasible expert which incurs zero constraint violation on every round, \emph{i.e.,} $g_t(i^\star)=0, \forall t.$  
The cumulative surrogate cost incurred by expert $i^\star$ can be upper bounded as:
\begin{eqnarray*}
	L_T (i^\star) = \sum_{t=1}^T \big(f_t(i^\star)+ \Phi'(Q(t)) g_t(i^\star)\big) \leq T,
\end{eqnarray*}
where we have used the fact that $||f_t||_\infty \leq 1$ and $g_t(i^\star) = 0, \forall t\geq 1.$ Hence, using the small-loss regret bound \eqref{adapt-hedge-reg-bd}, the regret for the surrogate cost functions with respect to any feasible arm $i^\star$ can be upper bounded as: 
\begin{eqnarray} \label{surr-reg-bd}
	\textrm{Regret}_T'(i^\star) \leq 
	  c\sqrt{T(1+\Phi'(Q(T))\ln N} + c(1+\Phi'(Q(T))) \ln N, 
\end{eqnarray}
where we have used the fact that $\max(2\gamma, 7\gamma^2) \leq 10 \equiv c$ (say). Using the inequality $\sqrt{x+y} \leq \sqrt{x}+\sqrt{y}, x \geq 0, y \geq 0,$ and substituting the upper bound from Eqn.\ \eqref{surr-reg-bd} into the regret decomposition inequality \eqref{reg-decomp-ineq}, we obtain
\begin{eqnarray} \label{reg-decomp-small-loss}
	\Phi(Q(T)) + \textrm{Regret}_T(i^\star) &\leq& \Phi(Q(0))+ c\bigg(\sqrt{T \ln N} + 
	  \sqrt{T\Phi'(Q(T)) \ln N} \nonumber\\
	  && +\Phi'(Q(T)) \ln N + \ln N \bigg). 
\end{eqnarray}
 
 
 %

\paragraph{1. Bounding the CCV:} 

Using the fact that $\textrm{Regret}_T(i^\star) \geq -L_T(i^\star) \geq -T, $ inequality \eqref{reg-decomp-small-loss} yields
\begin{eqnarray} \label{ineq-reg}
e^{\lambda Q(T)} \leq T + 1+ c\sqrt{T \ln N} + c\sqrt{\lambda T e^{\lambda Q(T)} \ln N } + (\lambda c \ln N) e^{\lambda Q(T)} + c \ln N.
\end{eqnarray}
Since $T\geq 1,$ our choice of the parameter $\lambda = T^{-(1-\beta)}/(2c\ln N)$ ensures that 
$\lambda c \ln N \leq  \nicefrac{1}{2}.$ 
Using this inequality to bound the coefficient of the penultimate term on the RHS of Eqn.\ \eqref{ineq-reg} and transposing, we obtain
\begin{eqnarray*}
	e^{\lambda Q(T)} \leq 8 \max \bigg(T+1, c \sqrt{T \ln N} ,c\sqrt{\lambda T \ln N}e^{\lambda Q(T)/2}, c\ln N\bigg). 
\end{eqnarray*} 
Since the maximum value on the RHS is achieved by at least one term on the right, comparing the LHS with each term on the RHS separately and simplifying,
we obtain the following bound for CCV:
\begin{eqnarray} \label{q-bd-new}
	Q(T) \leq \lambda^{-1}(c_1+ c_2\ln T + c_3 \ln \ln N),
\end{eqnarray}
where $c_1,c_2, c_3$ are universal constants. 
\paragraph{2. Bounding the Regret:}
Transposing the term $\Phi(Q(T))= e^{\lambda Q(T)}$ to the RHS of Eqn.\ \eqref{reg-decomp-small-loss} and using the fact that $\lambda c \ln N \leq  \nicefrac{1}{2}$ for our chosen parameter $\lambda,$ we obtain 
\begin{eqnarray*}
	\textrm{Regret}_T(i^\star) \leq 1+ c\sqrt{T \ln N} + c\sqrt{\lambda T \ln N}e^{\lambda Q(T)/2} - \frac{1}{2}e^{\lambda Q(T)}  + c \ln N. 
\end{eqnarray*}
Using the fact that $ax-bx^2 \leq \frac{a^2}{4b}, \forall b>0$,  the above inequality implies the following regret bound:
\begin{eqnarray} \label{reg-bd-new}
		\textrm{Regret}_T(i^\star) \leq 1+ c\sqrt{T \ln N} + \frac{c^2}{2} \lambda T \ln N +c \ln N.
\end{eqnarray} 
Substituting $\lambda = \frac{T^{-(1-\beta)}}{2c\ln N}$ into Eqns.\ \eqref{reg-bd-new} and Eqn.\ \eqref{q-bd-new}, we obtain the following regret and CCV bounds: 
\begin{eqnarray*} \label{reg-ccv-tradeoff}
	\textrm{Regret}_T(i^\star) = O(\sqrt{T\ln N}+ T^\beta +\ln N), ~ \textrm{CCV}_T =O(T^{1-\beta}\ln N \ln T).
\end{eqnarray*}
\hfill $\square$

\section{Convex Cost and Constraint functions} \label{gen_set}
In this section, we generalize our previous results to the general convex setting. In particular, we make the following assumption.
\begin{assumption}[Convexity and Lipschitzness] \label{lip-assump}
	All cost and constraint functions are convex and $G$-Lipschitz. The decision set $\mathcal{X}$ is a bounded subset of the $d$-dimensional Euclidean space $\mathbb{R}^d$.
\end{assumption}
We begin by recalling the notion of a $\delta$-cover from \citet{wainwright2019high}. See Figure \ref{delta-cover} in the Appendix for a schematic.
\begin{definition}[Covering number]
	A $\delta$-cover of a set $\mathbb{T}$ with respect to a metric $\rho$ is a set $\{\theta^1, \theta^2, \ldots, \theta^N \} \subseteq \mathbb{T}$ such that for each $\theta \in \mathbb{T},$ there exists some $i \in [N]$ such that $\rho(\theta, \theta^i) \leq \delta.$ The $\delta$-covering number $N(\delta; \mathbb{T}, \rho)$ is the cardinality of the smallest $\delta$-cover.  
\end{definition}
\paragraph{Construction:} Let $\mathcal{N}_\delta = \{x^1, x^2, \ldots, x^{N_\delta} \}$ be the smallest $\delta$-cover of the decision set $\mathcal{X}$ with $\delta = \nicefrac{1}{T}.$
Since $\mathcal{X}$ is contained within a $d$-dimensional ball of diameter $D,$ its covering number is bounded by $N_\delta \leq (1+ \frac{2D}{\delta})^d$ \citep[Lemma 5.7]{wainwright2019high}. We construct an instance of the \textsc{Constrained Expert} problem with $N_\delta$ experts where the $i$\textsuperscript{th} expert corresponds to the point $x^i, i \in [N_\delta].$ The cost and constraint violation for the experts are defined as:
 \begin{eqnarray} \label{expert-cost-constr}
f_t^{\texttt{CE}}(i) = f_t(x^i), ~~ g_t^{\texttt{CE}}(i) = (g_t(x^i) - G\delta)^+, ~ i \in [N_\delta].
\end{eqnarray}
Intuitively, the learner reduces the complex continuous decision space to a finite set of representative points, allowing us to apply the \textsc{Constrained Expert} algorithm.
\paragraph{Feasibility:} To show the feasibility of the above \textsc{Constrained Expert} problem, consider any feasible action in the decision set $x^\star \in \mathcal{X}$ that satisfies $g_t(x^\star)=0, \forall t.$ Let $x^{i^\star}$ be the nearest point in the $\delta$-cover, such that $||x^{i^\star}-x^\star|| \leq \delta$. Using the Lipschitzness of the constraint function, we have \[g_t^{\texttt{CE}}(i^\star) \equiv g_t(x^{i^\star}) \leq g_t(x^\star) + G||x^{i^\star}-x^\star|| \leq 0+ G\delta.\] 
Hence, from Eqn.\ \eqref{expert-cost-constr}, it follows that expert $i^\star$ is feasible as $g_t^{\texttt{CE}}(i^\star) =0, \forall t.$
\paragraph{Algorithm:} Our policy is summarized in Algorithm \ref{coco_alg2} where, on every round, we run Algorithm \ref{coco_alg} on the \textsc{Constrained Expert} instance defined above. We then use the output distribution from Algorithm \ref{coco_alg} to select the next action as the corresponding convex combination of the points in $\mathcal{N}_\delta$. 
\begin{algorithm}[htb]
   \caption{COCO Algorithm for Convex Cost and Constraints}
   \label{coco_alg2}
\begin{algorithmic}[1]
 \State {\bfseries Input:} A minimal $\delta$-cover of $\mathcal{X}$: $\mathcal{N}_\delta = \{x^1, x^2, \ldots, x^{N_\delta} \}$, with $\delta = \nicefrac{1}{T}.$ Parameter $\beta \in [0,1].$
	   \ForEach{$t=1:T$}
	   \State Run Algorithm \ref{coco_alg} on the  \textsc{Constrained Expert} problem with $N_\delta$ experts where the cost and constraint vectors are given by Eqn.\ \eqref{expert-cost-constr}. Obtain the distribution $p_t$ over $\mathcal{N}_\delta.$
	   \State Play $x_t = \sum_{i} p_t(i)x^i$
	   \EndFor
\end{algorithmic}
\end{algorithm}
\paragraph{Analysis:} 
 By Eqn.\ \eqref{expert-cost-constr} and Jensen's inequality, the cost and constraint violation on any round can be upper bounded by the cost and constraint violation of the \textsc{Constrained Expert} instance as: 
\begin{eqnarray*}
	f_t(x_t) = f_t (\sum_{i} p_t(i)x^i) \leq \sum_i p_t(i)f_t(x^i) = \langle f_t^{\texttt{CE}}, p_t \rangle, ~
	g_t(x_t) = g_t (\sum_{i} p_t(i)x^i) \leq \langle g_t^{\texttt{CE}},p_t \rangle + G\delta. 
\end{eqnarray*}
In addition, using the Lipschitzness of $f_t,$ we have $f_t(x^{i^\star}) - f_t(x^\star) \leq G\delta, \forall t. $ Hence, the regret and CCV of Algorithm \ref{coco_alg2} differs from that of the \textsc{Constrained Expert} instance by at most $G\delta T \leq G,$ which is a constant. 
Finally, using the fact that $\ln N_\delta = O(d \ln T),$ we invoke Theorem \ref{main-th} to obtain the following bounds for Algorithm \ref{coco_alg2}. The results are summarized in Theorem \ref{cvx-res}. 
\begin{eqnarray} \label{gen-prob-bd}
	\textrm{Regret}_T(x^\star) = O(\sqrt{dT\ln (T)}+ T^\beta +d\ln T ), ~ \textrm{CCV}_T =O(dT^{1-\beta} (\ln T)^2).
\end{eqnarray}
\begin{theorem} \label{cvx-res}
	Under Assumptions \ref{bdd-assump}, \ref{feasibility}, and \ref{lip-assump}, Algorithm \ref{coco_alg2} achieves  $\tilde{O}(\sqrt{dT}+ T^\beta)$ regret and $\tilde{O}(dT^{1-\beta})$ CCV for any $\beta \in [0,1]$.
\end{theorem}


\section{Convex and Smooth Cost and Constraint Functions} \label{cvx_case}
While the previous reduction-based approach, given by Algorithm \ref{coco_alg2}, is interesting, it can be computationally prohibitive when the dimension $d$ is large.  
To address this problem, we now propose an efficient gradient-based policy for the class of non-negative, smooth, and convex cost and constraint functions. We will make use of the following small-loss regret bound achieved by the Online Gradient Descent (OGD) policy for this class of functions. 
\begin{theorem}[\citet{orabona2019modern}, Theorem 4.25] \label{ogd-smooth}
	Let $\mathcal{X}$ be a closed non-empty convex decision set with diameter $D$. Let $l_1, l_2, \ldots, l_T$ be an arbitrary sequence of non-negative convex and $M$-smooth functions. Let $\nabla_t$ be the gradient of $l_t$ at $x_t, t \geq 1.$  Pick any $x_1 \in \mathcal{X},$ set the step sizes adaptively as $\eta_t = D/\sqrt{2\sum_{\tau=1}^t || \nabla_\tau||_2^2}, t\geq 1,$ and consider the Online Gradient Descent (OGD) policy with adaptive step sizes which selects the next action as:
		$x_{t+1} = \textsc{Proj}_{\mathcal{X}}(x_t - \eta_t \nabla_t),$ where
	$\textsc{Proj}_{\mathcal{X}}(\cdot)$ denotes the Euclidean projection operator on to the decision set $\mathcal{X}.$ Then we have:
	\begin{eqnarray} \label{smooth-reg}
		\textrm{Regret}_T(u) = \sum_{t=1}^T l_t(x_t) - \sum_{t=1}^T l_t(u) \leq 4D \sqrt{M \sum_{t=1}^T l_t(u)}+ 4D^2M, ~~u \in \mathcal{X}.
	\end{eqnarray}
\end{theorem}
In this section, we make the following assumption. 
\begin{assumption}[Convexity and Smoothness] \label{cvx-smooth}
	All cost and constraint functions are convex and $M$-smooth. 
\end{assumption}
\paragraph{Algorithm and Analysis:} Let $\Phi: \mathbb{R} \mapsto \mathbb{R}$ be a non-decreasing convex Lyapunov function, $x^\star \in \mathcal{X}$ be a feasible action, and $Q(t)$ be the CCV up to round $t$. Following identical arguments as in the \textsc{Constrained Expert} problem in Section \ref{simplex}, we define the surrogate cost function
\begin{eqnarray} \label{surr-cost-gen}
	\hat{f}_t(x) \coloneqq f_t(x) + \Phi'(Q(t))g_t(x), ~x \in \mathcal{X}.
\end{eqnarray} 
From Assumption \ref{cvx-smooth}, it follows that all surrogate cost functions are non-negative, convex, and $M_T \equiv M(1+\Phi'(Q(T))$-smooth. 
\begin{algorithm}[htb]
   \caption{COCO Algorithm for Smooth and Convex Cost and Constraints}
   \label{coco_alg3}
\begin{algorithmic}[1]
\State {\textbf{Initialization:} Choose $x_1 \in \mathcal{X}$ arbitrarily}
	   \ForEach{$t=1:T$}
	   \State Compute gradient $\nabla_t = \nabla \hat{f}_t(x_t)$ from Eqn.\ \eqref{surr-cost-gen} 
	   \State Set the step size $\eta_t = D/ \sqrt{2\sum_{\tau=1}^t || \nabla_\tau||_2^2}.$ 
	   \State Choose next action using OGD: $x_{t+1} = \textsc{Proj}_{\mathcal{X}}(x_t - \eta_t \nabla_t)$
	   \EndFor
\end{algorithmic}
\end{algorithm}
Our proposed algorithm is described in Algorithm \ref{coco_alg3} where we run the standard OGD policy on the surrogate cost functions $\{l_t \equiv \hat{f}_t\}_{t \geq 1},$ with adaptive step sizes given by Theorem \ref{ogd-smooth}. It is crucial to note that the step sizes $\{\eta_t\}_{t \geq 1}$, which are derived from the past gradients, are oblivious to the smoothness parameter $M_T,$ which is unknown \emph{a priori}. Using Eqn.\ \eqref{smooth-reg} from Theorem \ref{ogd-smooth}, the regret for the surrogate costs 
for any feasible action $x^\star$ can be bounded as:
\begin{eqnarray} \label{smooth-reg-decomp}
\textrm{Regret}_T'(x^\star)\leq   4D\sqrt{MT(1+\Phi'(Q(T))} + 4(1+\Phi'(Q(T))D^2M.
\end{eqnarray}
We have used the fact that the cumulative surrogate cost is upper bounded by $\sum_{t=1}^T \hat{f}_t(x^\star) \leq T$ as $g_t(x^\star)=0, \forall t \geq 1.$
The regret bound \eqref{smooth-reg-decomp} becomes algebraically identical to Eqn.\ \eqref{surr-reg-bd} under the substitutions $c \gets 4, \ln N \gets D^2M$. Thus, reusing the same analysis and parameter choices (including the Lyapunov function) from the proof of Theorem \ref{main-th}, we arrive at the following result.

\begin{theorem} \label{reg-ccv-smooth}
	Let $\beta \in [0,1]$ be a tunable parameter, and define the Lyapunov function as $\Phi(x)= e^{\lambda x}$ with $\lambda = T^{-(1-\beta)}/(8D^2M).$ Then, under Assumptions \ref{bdd-assump}, \ref{feasibility}, and \ref{cvx-smooth}, Algorithm \ref{coco_alg3} achieves the following guarantee for any feasible action $x^\star \in \mathcal{X}^\star$:
	\[\textrm{Regret}_T(x^\star) = O(D\sqrt{MT}+ T^\beta +D^2M), ~ \textrm{CCV}_T =O(T^{1-\beta}M \ln T).\]
	In particular, if $f_t=0, \forall t,$ upon setting $\beta=1,$ we obtain $\textrm{CCV}_T= O(M\ln T).$
	\end{theorem}
See Section \ref{relaxed_feasibility} in the Appendix for an extension of the above result that relaxes the feasibility assumption (Assumption \ref{feasibility}) by allowing the benchmark $x^\star$ to violate the constraints within a prescribed long term budget of $B_T.$

\section{Conclusion} \label{concl}
In this paper we propose online policies for the COCO problem that achieve improved cumulative constraint violation (CCV) by carefully trading it off with regret. These results are particularly important in applications where violating the constraints are costly, such as autonomous driving or budget-constrained advertising. An important direction for future work is to design computationally efficient algorithms that achieve sharper CCV guarantees in the fixed-dimensional setting without the smoothness assumption. Additionally, it would be interesting to establish matching lower bounds.

 \section{Acknowledgement}
This work was supported by the Department of Atomic Energy, Government of India, under project no. RTI4001 and by a Google India faculty research award. The authors gratefully acknowledge comments from the anonymous reviewers, which substantially improved the quality of the presentation. 


\clearpage
\bibliography{OCO.bib}

@article{supantha2025universal,
  title={Universal Dynamic Regret and Constraint Violation Bounds for Constrained Online Convex Optimization},
  author={Supantha, Subhamon and Sinha, Abhishek},
  journal={arXiv preprint arXiv:2510.01867},
  year={2025}
}

@article{sarkar2025online,
  title={Online Learning for Approximately-Convex Functions with Long-term Adversarial Constraints},
  author={Sarkar, Dhruv and Mukhopadhyay, Samrat and Sinha, Abhishek},
  journal={arXiv preprint arXiv:2508.16992},
  year={2025}
}

@inproceedings{mhammedi2019lipschitz,
  title={Lipschitz adaptivity with multiple learning rates in online learning},
  author={Mhammedi, Zakaria and Koolen, Wouter M and Van Erven, Tim},
  booktitle={Conference on Learning Theory},
  pages={2490--2511},
  year={2019},
  organization={PMLR}
}

@article{cesa2007improved,
  title={Improved second-order bounds for prediction with expert advice},
  author={Cesa-Bianchi, Nicolo and Mansour, Yishay and Stoltz, Gilles},
  journal={Machine Learning},
  volume={66},
  number={2},
  pages={321--352},
  year={2007},
  publisher={Springer}
}

@article{amodei2016concrete,
  title={Concrete problems in AI safety},
  author={Amodei, Dario and Olah, Chris and Steinhardt, Jacob and Christiano, Paul and Schulman, John and Man{\'e}, Dan},
  journal={arXiv preprint arXiv:1606.06565},
  year={2016}
}

@inproceedings{koolen2015second,
  title={Second-order quantile methods for experts and combinatorial games},
  author={Koolen, Wouter M and Van Erven, Tim},
  booktitle={Conference on Learning Theory},
  pages={1155--1175},
  year={2015},
  organization={PMLR}
}

@article{lekeufack2024optimistic,
  title={An Optimistic Algorithm for Online Convex Optimization with Adversarial Constraints},
  author={Lekeufack, Jordan and Jordan, Michael I},
  journal={arXiv preprint arXiv:2412.08060},
  year={2024}
}

@article{lu2025order,
  title={Order-Optimal Projection-Free Algorithm for Adversarially Constrained Online Convex Optimization},
  author={Lu, Yiyang and Pedramfar, Mohammad and Aggarwal, Vaneet},
  journal={arXiv preprint arXiv:2502.16744},
  year={2025}
}

@article{cesa1997use,
  title={How to use expert advice},
  author={Cesa-Bianchi, Nicolo and Freund, Yoav and Haussler, David and Helmbold, David P and Schapire, Robert E and Warmuth, Manfred K},
  journal={Journal of the ACM (JACM)},
  volume={44},
  number={3},
  pages={427--485},
  year={1997},
  publisher={ACM New York, NY, USA}
}

@article{auer2002adaptive,
  title={Adaptive and self-confident on-line learning algorithms},
  author={Auer, Peter and Cesa-Bianchi, Nicolo and Gentile, Claudio},
  journal={Journal of Computer and System Sciences},
  volume={64},
  number={1},
  pages={48--75},
  year={2002},
  publisher={Elsevier}
}

@article{hazan2010extracting,
  title={Extracting certainty from uncertainty: Regret bounded by variation in costs},
  author={Hazan, Elad and Kale, Satyen},
  journal={Machine learning},
  volume={80},
  pages={165--188},
  year={2010},
  publisher={Springer}
}

@article{lecnotes_luo,
  title={Lecture 4},
  author={Luo, Haipeng},
  journal={Introduction to Online Learning},
  note={\url{https://haipeng-luo.net/courses/CSCI699/index.html}},
  year={2017}
}

@article{vaze2025sqrt,
  title={$ O (\sqrt{T}$)  Static Regret and Instance Dependent Constraint Violation for Constrained Online Convex Optimization},
  author={Vaze, Rahul and Sinha, Abhishek},
  journal={arXiv preprint arXiv:2502.05019},
  year={2025}
}

@article{yu2020low,
  title={A low complexity algorithm with  $\mathcal{O}(\sqrt{T})$ regret and $\mathcal{O}(1)$ constraint violations for online convex optimization with long term constraints},
  author={Yu, Hao and Neely, Michael J},
  journal={Journal of Machine Learning Research},
  volume={21},
  number={1},
  pages={1--24},
  year={2020}
}

@article{wang2025revisiting,
  title={Revisiting Projection-Free Online Learning with Time-Varying Constraints},
  author={Wang, Yibo and Wan, Yuanyu and Zhang, Lijun},
  journal={arXiv preprint arXiv:2501.16046},
  year={2025}
}

@inproceedings{
sinha2024optimal,
title={Optimal Algorithms for Online Convex Optimization with Adversarial Constraints},
author={Abhishek Sinha and Rahul Vaze},
booktitle={The Thirty-eighth Annual Conference on Neural Information Processing Systems},
year={2024},
url={https://openreview.net/forum?id=TxffvJMnBy}
}

@inproceedings{yi2021regret,
  title={Regret and cumulative constraint violation analysis for online convex optimization with long term constraints},
  author={Yi, Xinlei and Li, Xiuxian and Yang, Tao and Xie, Lihua and Chai, Tianyou and Johansson, Karl},
  booktitle={International Conference on Machine Learning},
  pages={11998--12008},
  year={2021},
  organization={PMLR}
}

@article{ruder2017overview,
  title={An overview of multi-task learning in deep neural networks},
  author={Ruder, Sebastian},
  journal={arXiv preprint arXiv:1706.05098},
  year={2017}
}

@article{guo2022online,
  title={Online convex optimization with hard constraints: Towards the best of two worlds and beyond},
  author={Guo, Hengquan and Liu, Xin and Wei, Honghao and Ying, Lei},
  journal={Advances in Neural Information Processing Systems},
  volume={35},
  pages={36426--36439},
  year={2022}
}

@inproceedings{dekel2006online,
  title={Online multitask learning},
  author={Dekel, Ofer and Long, Philip M and Singer, Yoram},
  booktitle={International Conference on Computational Learning Theory},
  pages={453--467},
  year={2006},
  organization={Springer}
}

@InProceedings{georgios-cautious,
  title = 	 {Cautious Regret Minimization: Online Optimization with Long-Term Budget Constraints},
  author =       {Liakopoulos, Nikolaos and Destounis, Apostolos and Paschos, Georgios and Spyropoulos, Thrasyvoulos and Mertikopoulos, Panayotis},
  booktitle = 	 {Proceedings of the 36th International Conference on Machine Learning},
  pages = 	 {3944--3952},
  year = 	 {2019},
  editor = 	 {Chaudhuri, Kamalika and Salakhutdinov, Ruslan},
  volume = 	 {97},
  series = 	 {Proceedings of Machine Learning Research},
  month = 	 {09--15 Jun},
  publisher =    {PMLR},
  url = 	 {https://proceedings.mlr.press/v97/liakopoulos19a.html}
}

@InProceedings{pmlr-v70-sun17a,
  title = 	 {Safety-Aware Algorithms for Adversarial Contextual Bandit},
  author =       {Wen Sun and Debadeepta Dey and Ashish Kapoor},
  booktitle = 	 {Proceedings of the 34th International Conference on Machine Learning},
  pages = 	 {3280--3288},
  year = 	 {2017},
  editor = 	 {Precup, Doina and Teh, Yee Whye},
  volume = 	 {70},
  series = 	 {Proceedings of Machine Learning Research},
  month = 	 {06--11 Aug},
  publisher =    {PMLR},
  url = 	 {https://proceedings.mlr.press/v70/sun17a.html}
}

@article{yi2023distributed,
  title={Distributed Online Convex Optimization with Adversarial Constraints: Reduced Cumulative Constraint Violation Bounds under Slater's Condition},
  author={Yi, Xinlei and Li, Xiuxian and Yang, Tao and Xie, Lihua and Hong, Yiguang and Chai, Tianyou and Johansson, Karl H},
  journal={arXiv preprint arXiv:2306.00149},
  year={2023}
}

@article{neely2017online,
  title={Online convex optimization with time-varying constraints},
  author={Neely, Michael J and Yu, Hao},
  journal={arXiv preprint arXiv:1702.04783},
  year={2017}
}

@article{mannor2009online,
  title={Online Learning with Sample Path Constraints.},
  author={Mannor, Shie and Tsitsiklis, John N and Yu, Jia Yuan},
  journal={Journal of Machine Learning Research},
  volume={10},
  number={3},
  year={2009}
}

@inproceedings{jenatton2016adaptive,
  title={Adaptive algorithms for online convex optimization with long-term constraints},
  author={Jenatton, Rodolphe and Huang, Jim and Archambeau, C{\'e}dric},
  booktitle={International Conference on Machine Learning},
  pages={402--411},
  year={2016},
  organization={PMLR}
}

@article{yuan2018online,
  title={Online convex optimization for cumulative constraints},
  author={Yuan, Jianjun and Lamperski, Andrew},
  journal={Advances in Neural Information Processing Systems},
  volume={31},
  year={2018}
}

@InProceedings{sinha2023banditq,
  title = 	 {Bandit{Q}:{F}air {B}andits with {G}uaranteed {R}ewards},
  author =       {Sinha, Abhishek},
  booktitle = 	 {Proceedings of the Fortieth Conference on Uncertainty in Artificial Intelligence},
  pages = 	 {3227--3244},
  year = 	 {2024},
  editor = 	 {Kiyavash, Negar and Mooij, Joris M.},
  volume = 	 {244},
  series = 	 {Proceedings of Machine Learning Research},
  month = 	 {15--19 Jul},
  publisher =    {PMLR},
  url = 	 {https://proceedings.mlr.press/v244/sinha24a.html}
}

@article{bubeck2012regret,
  title={Regret analysis of stochastic and nonstochastic multi-armed bandit problems},
  author={Bubeck, S{\'e}bastien and Cesa-Bianchi, Nicolo and others},
  journal={Foundations and Trends{\textregistered} in Machine Learning},
  volume={5},
  number={1},
  pages={1--122},
  year={2012},
  publisher={Now Publishers, Inc.}
}

@book{cesa2006prediction,
  title={Prediction, learning, and games},
  author={Cesa-Bianchi, Nicolo and Lugosi, G{\'a}bor},
  year={2006},
  publisher={Cambridge university press}
}

@book{hazan2022introduction,
  title={Introduction to online convex optimization},
  author={Hazan, Elad},
  year={2022},
  publisher={MIT Press}
}

@misc{coco-code-improved,
  author = {Abhishek Sinha},
  title = {Source Code for ``{B}eyond $\tilde{O}(\sqrt{T})$ {C}onstraint {V}iolation for {O}nline {C}onvex {O}ptimization with {A}dversarial {C}onstraints"; {A.} {S}inha, {R.} {V}aze},
  year = {2025},
  publisher = {GitHub},
  journal = {GitHub repository},
  howpublished = {\url{https://github.com/abhishek-sinha-tifr/COCO-improved-CCV}},
  commit = {9c5e5d1}
}

@article{orabona2019modern,
  title={A modern introduction to online learning},
  author={Orabona, Francesco},
  journal={arXiv preprint arXiv:1912.13213},
  year={2019}
}

@book{wainwright2019high,
  title={High-dimensional statistics: A non-asymptotic viewpoint},
  author={Wainwright, Martin J},
  volume={48},
  year={2019},
  publisher={Cambridge University Press}
}
\bibliographystyle{unsrtnat}
\clearpage
\section{Appendix} \label{app}

\subsection{On the Boundedness Assumption} \label{bdd}
Assumption \ref{bdd-assump} implies that there exist  constants $K_f, K_g$ such that 
\[ |f_t (x)| \leq K_f, ~~|g_t(x)| \leq K_g, \forall x \in \mathcal{X}, \forall t \geq 1. \] 
Since the constraint function $g_t$ is non-negative, the above implies 
\[ |f_t (x)| \leq K_f, ~~0\leq g_t(x) \leq K_g, \forall x \in \mathcal{X}, \forall t \geq 1. \]
Now consider the following translated and scaled version of the cost and constraint functions:
\[\tilde{f}_t (x)  \coloneqq \frac{f_t(x)}{2K_f} + \frac{1}{2}, ~~ \tilde{g}_t(x) \coloneqq \frac{g_t(x)}{K_g}, x \in \mathcal{X}, t\geq 1.\]
It is easy to verify that $0 \leq \tilde{f}_t, \tilde{g}_t \leq 1, \forall t\geq 1.$ Hence, we can work with these modified cost and constraint functions. 

\textbf{P.S.} With the feasibility assumption (Assumption \ref{feasibility}), we can even obtain an explicit expression for $K_g$. Let $x^\star$ be a feasible action. Using the $G$-Lipschitzness of the function $g_t,$ we have 
\[g_t(x) \leq g_t(x^\star) + G||x_t - x^\star|| \leq 0 + GD.\]
Thus we can take $K_g \equiv GD.$ Consequently, we only need to assume the cost functions to be bounded. 

\subsection{Pseudocode for the Adaptive Hedge Policy} \label{adapt-hedge-pseudocode}
\begin{algorithm}[htb]
   \caption{Adaptive Hedge Algorithm for the \textsc{Expert} problem with Unbounded Losses}
   \label{adapt_hedge}
\begin{algorithmic}[1]
   \State {\bfseries Input:} Number of experts $N,$ Horizon length $T$, Non-negative and potentially unbounded loss vectors $\{l_t\}_{t=1}^T, \gamma \geq 1.$ 
  
  \State {\bfseries Initialization:} Algorithm's loss $\tilde{L}_0=0$, Losses of the experts $L_0 = \bm{0},$ $G_0=1$.
   \ForEach{$t=1:T$}
   \State Compute self-confident learning rate
        \begin{eqnarray} \label{learning-rate}
 	\eta_t = \frac{1}{\sqrt{G_{t-1}}}\sqrt{\frac{\ln N}{\tilde{L}_{t-1}+ \gamma G_{t-1}}}.
 \end{eqnarray} 
    \State Play adaptive Hedge by choosing the $i$\textsuperscript{th} expert with probability
    \begin{eqnarray*}
    	    p_t(i) = \exp(-\eta_t L_{t-1}(i))/\sum_{j=1}^N \exp(-\eta_t L_{t-1}(j)), ~\forall i \in [N].
    \end{eqnarray*}
     \State The loss vector $l_t$ is revealed to the learner
        \State Update the algorithm's and experts' losses
        \begin{eqnarray*}
        	        \tilde{L}_t \gets \tilde{L}_{t-1} + \langle l_t, p_t\rangle, ~L_t \gets L_{t-1} + l_t
        \end{eqnarray*}
    \State $G_t$ is an upper bound to $||l_t||_\infty$ such that $ 1 \leq G_{t+1}/G_{t} \leq \gamma $
  \EndForEach
\end{algorithmic}
\end{algorithm}

\subsection{Proof of Theorem \ref{gen-small-loss} } \label{gen-small-loss-proof}
Our proof adapts the arguments given in \citet{lecnotes_luo} while accounting for unbounded losses. 
	Let $L_t(i)$ denote the cumulative loss of the $i$\textsuperscript{th} expert up to round $t$, \emph{i.e.,} $L_t(i)= \sum_{\tau=1}^t l_\tau(i), i\in [N]$. Define the potential function $\Phi_t(\eta) = \frac{1}{\eta} \log \left( \frac{1}{N} \sum_{i=1}^{N} \exp(-\eta L_t(i)) \right)$. We have 
	\begin{eqnarray*}
		\Phi_t (\eta_t) - \Phi_{t-1}(\eta_t) &=& \frac{1}{\eta_t} \ln \bigg( \frac{\sum_{i=1}^N \exp(-\eta_t L_t(i))}{\sum_{i=1}^N \exp(-\eta_t L_{t-1}(i))}\bigg) \\
		&=& \frac{1}{\eta_t} \ln \bigg(\sum_{i=1}^N p_t(i)\exp(-\eta_t l_t(i)) \bigg) \\
		&\stackrel{(a)}{\leq} & \frac{1}{\eta_t} \ln \bigg(\sum_{i=1}^N p_t(i) \big(1 -\eta_t l_t(i) + \frac{\eta_t^2 l_t^2(i)}{2} \big)\bigg) \\
		&=& \frac{1}{\eta_t} \ln \bigg(1 - \eta_t \langle p_t, l_t \rangle + \frac{\eta_t^2}{2} \sum_{i=1}^N p_t(i) l_t^2(i) \bigg) \\
		&\stackrel{(b)}{\leq} & -\langle p_t, l_t \rangle + \frac{\eta_t}{2} \sum_{i=1}^N p_t(i) l_t^2(i).
	\end{eqnarray*}
	where in inequality (a), we have used the fact that $e^{-y} \leq 1 - y + \frac{y^2}{2}, \forall  y \geq 0,$ and in (b), we have used the fact that $1+y \leq e^y, \forall y \in \mathbb{R}.$
Thus we have that
\begin{equation} \label{eq1t}
    \langle p_t, \ell_t \rangle  \leq  \Phi_{t-1}(\eta_t) - \Phi_t(\eta_t) + \frac{\eta_t}{2} \sum_{i=1}^{N} p_t(i) \ell_t^2(i).
\end{equation}

Summing the above inequality for $1\leq t \leq T$ yields
\begin{eqnarray} \label{bd1}
    && \tilde{L}_T \nonumber \\
    &=&\sum_{t=1}^T \langle p_t, l_t \rangle \nonumber \\
    &\leq& \Phi_0(\eta_1) - \Phi_T(\eta_{T+1}) + \sum_{t=1}^{T} \frac{\eta_t}{2} \sum_{i=1}^{N} p_t(i) \ell_t^2(i) + \sum_{t=1}^{T} (\Phi_t(\eta_{t+1}) - \Phi_t(\eta_t)) \nonumber \\
    &\stackrel{(a)}{\leq}& \frac{\ln N}{\eta_{T+1}} - \frac{1}{\eta_{T+1}} \ln \left(\exp(-\eta_{T+1} L_T(i^*)) \right) + \sum_{t=1}^{T} \frac{\eta_t}{2} G_{t}\sum_{i=1}^{N} p_t(i) \ell_t(i) + \sum_{t=1}^{T} (\Phi_t(\eta_{t+1}) - \Phi_t(\eta_t))\nonumber \\
    &\stackrel{(b)}{\leq} & \sqrt{(\tilde{L}_{T} + \gamma G_{T}) G_{T}\ln N } + L_T(i^*) + \underbrace{ \gamma \sqrt{G_T\ln N}\sum_{t=1}^{T} \frac{\langle p_t, \ell_t\rangle}{2\sqrt{\tilde{L}_{t-1}+ \gamma G_{t-1}}}}_{(A)}
    +\underbrace{\sum_{t=1}^{T} (\Phi_t(\eta_{t+1}) - \Phi_t(\eta_t))}_{(B)}, \nonumber \\
\end{eqnarray}
where, in (a) we have used the fact that $||l_t||_\infty \leq G_t,$ and in (b) we have used the expression \eqref{learning-rate} for the learning rate $\eta_t$ along with the fact that $G_t \leq \gamma G_{t-1}$ and $G_t \leq G_T, \forall t.$ 

To bound term (A) in Eqn.\ \eqref{bd1}, observe that 
\[ \tilde{L}_t = \tilde{L}_{t-1} + \langle p_t, l_t \rangle \leq \tilde{L}_{t-1} + G_t \leq \tilde{L}_{t-1} + \gamma G_{t-1}.\] Hence,
\begin{eqnarray*}
    \sum_{t=1}^{T} \frac{\langle p_t, \ell_t \rangle}{\sqrt{\tilde{L}_{t-1} + \gamma G_{t-1}}} \leq  \sum_{t=1}^{T} \frac{\tilde{L}_t - \tilde{L}_{t-1}}{\sqrt{\tilde{L}_{t}}} 
    \stackrel{(c)}{\leq} \int_{\tilde{L}_0}^{\tilde{L}_T} \frac{dx}{\sqrt{x}}  
    = 2 \sqrt{\tilde{L}_T} ,
\end{eqnarray*}
where in inequality (c), we have used the monotonicity of the sequence $\{\tilde{L}_t\}_{t \geq 1}.$ 

Furthermore, it can be readily verified that the potential function $\Phi_t(\eta)$ is non-decreasing in $\eta $ as $\Phi_t'(\eta) \geq 0$ \citep{lecnotes_luo}. Since the learning rate $\eta_t$ is non-increasing in $t,$ we conclude that each term in term (B) in Eqn.\ \eqref{bd1} is non-positive.  
Hence, using the inequality $\sqrt{a+b} \leq \sqrt{a} + \sqrt{b},$ from Eqn.\ \eqref{bd1}, we have
\begin{eqnarray*}
	\tilde{L}_T \leq 2\gamma \sqrt{\tilde{L}_{T} G_T \ln N } + L_T(i^*)   + G_T\sqrt{\gamma \ln N }.
\end{eqnarray*}
In the above, we have used the fact that $\gamma \geq 1.$
Solving the above quadratic inequality using Lemma \ref{quadratic} below, we conclude that the adaptive Hedge policy enjoys the following small-loss regret bound for the \textsc{Expert} problem:
\begin{eqnarray*}
	\tilde{L}_T - L_T(i^\star) \leq 2\gamma \sqrt{L_T(i^\star) G_T \ln N } +  7 \gamma^2 G_T \ln N.
\end{eqnarray*}

\subsection{Proof of Auxiliary Lemmas}

\begin{lemma}[A quadratic inequality] \label{quadratic}
	Consider the following quadratic inequality $x^2 \leq ax +b$, where $a\geq 0, b\geq 0.$ Then we have $x^2 \leq a^2 + b + a\sqrt{b}.$
\end{lemma}
\begin{proof}
	Solving the given inequality using the usual quadratic formula and the fact that $\sqrt{x+y} \leq \sqrt{x} + \sqrt{y}, ~x, y \geq 0,$, we have 
	\begin{eqnarray*}
		x \leq \frac{a+\sqrt{a^2+4b}}{2} \leq \frac{a+a+2\sqrt{b}}{2} = a + \sqrt{b}.
	\end{eqnarray*}
	Hence,
	\begin{eqnarray*}
		x^2 \leq a(a+\sqrt{b}) + b = a^2 + b + a\sqrt{b}.
 	\end{eqnarray*}
\end{proof}

	\begin{lemma}[Bounding $\gamma$] \label{ratio-bd}
Let $\Phi(x)=e^{\lambda x}, \lambda = T^{-(1-\beta)}/(2c\ln N)$, where $c=10.$ Define $G_t=1+\Phi'(Q(t)), t \geq 0$ and $\gamma \equiv \max_{1 \leq t \leq T} \frac{G_t}{G_{t-1}}$. Then we have
	$ \gamma \leq 1.08. $ 
\end{lemma}

\begin{proof}
	Since $0\leq g_t \leq 1,$ from Eqn.\ \eqref{q-ev3}, we have 
\begin{eqnarray} \label{q-inc-bd}
	Q(t) = Q(t-1)+ \langle g_t, p_t\rangle \leq Q(t-1)+1.
\end{eqnarray}
Thus
\begin{eqnarray} \label{ratio-bd-eq}
    \frac{G_t}{G_{t-1}} &=&  \frac{1 + \lambda e^{\lambda Q(t)}}{1 + \lambda e^{\lambda Q(t-1)}} \stackrel{(a)}{\leq} \frac{1 + \lambda e^\lambda e^{\lambda Q(t-1)}}{1 + \lambda e^{\lambda Q(t-1)}} \leq e^\lambda. 
    \end{eqnarray}
    where in (a) we have used inequality \eqref{q-inc-bd}. Since $N\geq 2$ and $T \geq 1,$ we have $\lambda \leq (2c \ln 2)^{-1}.$ Hence, Eqn.\ \eqref{ratio-bd-eq} implies that 
    \[\gamma \leq e^{1/(20\ln 2)} \leq 1.08.\]
\end{proof}

\subsection{Construction of a Minimal $\delta$-cover of the Decision Set $\mathcal{X}$}
\begin{figure*}[h]
\centering
	\includegraphics[scale=1.2]{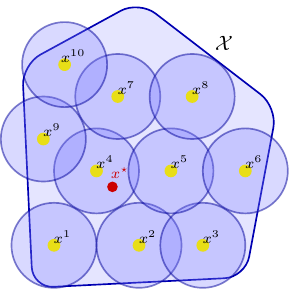}
	\caption{Schematic depicting the greedy construction of a minimal $\delta$-cover of the decision set $\mathcal{X}$. If a point $x \in \mathcal{X}$ is not covered yet, we construct a ball of radius $\delta$ centred at $x$ and include the point $x$ in the $\delta$-cover. The process continues till the entire set $\mathcal{X}$ is covered. }
	\label{delta-cover}
\end{figure*}

\subsection{Relaxing the Feasibility Assumption} \label{relaxed_feasibility}
In this Section, we relax the feasibility Assumption \ref{feasibility} by allowing the benchmark actions to violate the constraints up to a given budget $B_T\geq 0$. In particular, we consider the following feasible set of actions which are budget feasible in the long term: 
\begin{eqnarray} \label{budget-constr}
	\mathcal{X}^\star_{B_T} = \{x \in \mathcal{X}: \sum_{t=1}^T g_t(x) \leq B_T \}. 
\end{eqnarray}
We now replace Assumption \ref{feasibility} with the following assumption.

\begin{assumption}[Long-term Feasibility] \label{feasibility-new}
	The feasible set is non-empty, \emph{i.e.,} $\mathcal{X}_{B_T}^\star \neq \emptyset.$ 
	\end{assumption}

Note that $\mathcal{X}^\star_{B_T} = \mathcal{X}^\star$ when $B_T=0.$ Since $0\leq g_t \leq 1$ (Assumption \ref{bdd-assump}), we clearly have $\mathcal{X}^\star_{T} = \mathcal{X},$ where $\mathcal{X}$ is the entire decision set. Thus, without any loss of generality, we can assume that $B_T \leq T.$ 
\subsubsection{The Generalized Regret Decomposition Inequality} 
We now generalize the regret decomposition inequality \eqref{reg-decomp-ineq} by taking into account the long-term feasibility constraint \eqref{budget-constr}. Let $\Phi: \mathbb{R} \mapsto \mathbb{R}$ be any non-decreasing convex Lyapunov function defined as in Section \ref{cvx_case} and let $x^\star \in \mathcal{X}^\star_{B_T}$ be any long-term feasible action satisfying the budget constraint \eqref{budget-constr}. Recall that CCV evolves as $Q(t)=Q(t-1)+g_t(x_t).$ Hence, using the convexity of $\Phi(\cdot)$, we obtain 
\begin{eqnarray*}
    &&\Phi(Q(t)) - \Phi(Q(t-1)) + \big(f_t(x_t) - f_t(x^\star)\big) \\
    &&\le \big(f_t(x_t) + \Phi'(Q(t)) g_t(x_t)\big)
    - \big(f_t(x^\star) + \Phi'(Q(t)) g_t(x^\star)\big)
    + \Phi'(Q(t)) g_t(x^\star).
\end{eqnarray*}
where, unlike in Section \ref{cvx_case}, the violation $g_t(x^\star)$ in this case could be strictly positive. Summing up the above inequalities, we conclude
\begin{eqnarray} \label{reg-decomp-gen-2}
    \Phi(Q(T)) - \Phi(Q(0)) + \text{Regret}_T(x^\star)
    &\stackrel{(a)}{\leq}& \text{Regret}_T'(x^\star) + \Phi'(Q(T)) \sum_{t=1}^T g_t(x^\star) \nonumber  \\
    &\stackrel{(b)}{\leq}& \text{Regret}_T'(x^\star) + \Phi'(Q(T)) B_T, 
\end{eqnarray}
where, as before, $\text{Regret}_T(x^\star)$ and $\text{Regret}_T'(x^\star)$ denote the regrets for learning the original cost functions $\{f_t\}_{t \geq 1}$ and the surrogate cost functions $\{\hat{f}_t\}_{t \geq 1}$ respectively w.r.t. the long-term feasible benchmark $x^\star$ (see Eqn. \eqref{surr-cost-gen} for the definition of the surrogate cost functions). 
In the above inequality, the upper bound in step (a) follows from the monotonicity of the CCV $(Q(t))_{t \ge 1}$ and the convexity of the Lyapunov function $\Phi(\cdot)$, while step (b) uses the budget constraint for $x^\star \in \mathcal{X}^\star_{B_T}$.

In the following, we consider the case of convex and smooth cost and constraint functions under the relaxed feasibility assumption (see Section \ref{cvx_case}). The analysis for the \textsc{Constrained Expert} problem is identical.

\subsubsection{Convex and Smooth Cost and Constraints with Long-term Budget Constraints}
Inequality \eqref{reg-decomp-gen-2} is of the same form as the regret decomposition inequality \eqref{reg-decomp-ineq} under Assumption \ref{feasibility}, albeit with an extra additive term $\Phi'(Q(T)) B_T$ appearing on the right-hand side. By using the same exponential Lyapunov function as before (with the parameter $\lambda$ now adapted to the budget $B_T$), we can analogously solve the above functional inequality and derive the corresponding regret and CCV bounds under long-term violation budget constraints. 

The cumulative cost of the surrogate functions under any long-term feasible action $x^\star \in \mathcal{X}^\star_{B_T}$ can be upper bounded as
\begin{eqnarray} \label{cum-loss-2}
	\sum_{t=1}^T \hat{f}_t(x^\star) = \sum_{t=1}^T f_t(x^\star) + \sum_{t=1}^T \Phi'(Q(t))g_t(x^\star) \stackrel{(a)}{\leq} T + \Phi'(Q(T))\sum_{t=1}^T g_t(x^\star) \stackrel{(b)}{\leq} T + \Phi'(Q(T))B_T,
\end{eqnarray}
where in inequality (a), we have again made use of the convexity of the Lyapunov function and the non-decreasing property of the CCV, and in (b), we have used the budget constraint for $x^\star$. Hence, using Eqn.\ \eqref{smooth-reg} from Theorem \ref{ogd-smooth}, the regret for the surrogate costs under the action of Algorithm \ref{coco_alg3} 
can be upper bounded as:
\begin{eqnarray} \label{smooth-reg-decomp2}
\textrm{Regret}_T'(x^\star)&\leq&   4D\sqrt{M(T+\Phi'(Q(T)B_T)(1+\Phi'(Q(T))} + 4(1+\Phi'(Q(T))D^2M \nonumber \\
&\leq & c_1 \sqrt{T} + c_2 \sqrt{T \Phi'(Q(T))} + c_3 \Phi'(Q(T)) \sqrt{B_T},
\end{eqnarray}
where $c_1, c_2, c_3$ are generic constants which depend only on the problem-specific parameters $M$ and $D$. If necessary, the reader can easily figure out the explicit values of these constants in each line of the derivation below. 

As in Section \ref{constr-sec}, we now set $\Phi(\cdot)$ to be the exponential Lyapunov function, \emph{i.e.,} $\Phi(x):= \exp(\lambda x),$ for a suitable value of the parameter $\lambda$ which will be fixed later. With this choice, the regret decomposition inequality \eqref{reg-decomp-gen-2} yields
\begin{eqnarray*} 
	e^{\lambda Q(T)} + \textrm{Regret}_T(x^\star) \leq c_1 \sqrt{T} + c_2 \sqrt{\lambda T} e^{\lambda Q(T)/2} + c_3 \lambda B_T e^{\lambda Q(T)}. 
\end{eqnarray*}
We now choose some $\lambda$ such that $\lambda \leq (2c_3 B_T)^{-1}.$ Hence, the above equation yields
\begin{eqnarray} \label{reg-dec-BT}
	\frac{1}{2}e^{\lambda Q(T)} + \textrm{Regret}_T(x^\star) \leq c_1 \sqrt{T} +  c_2 \sqrt{\lambda T} e^{\lambda Q(T)/2}.
\end{eqnarray}
The Regret and CCV bounds are obtained by solving the above inequality. 
\paragraph{Bounding the CCV:}
Using the fact that $\textrm{Regret}_T(x^\star) \geq -T,$ Eqn.\ \eqref{reg-dec-BT} yields
\begin{eqnarray*}
	e^{\lambda Q(T)} \leq c_1T+ c_2 \sqrt{\lambda T} e^{\lambda Q(T)/2} \leq 2 \max(c_1T, c_2 \sqrt{\lambda T} e^{\lambda Q(T)/2}).
\end{eqnarray*}
This implies the following bounds for $Q(T):$
\begin{eqnarray*}
	Q(T) \leq \lambda^{-1}(c_1+ \log \lambda + \log T) = O(\lambda^{-1} \log T). 
\end{eqnarray*}

\paragraph{Bounding the Regret:} Starting from inequality \eqref{reg-dec-BT} once again, we have the following upper bound on regret
\begin{eqnarray*}
	\textrm{Regret}_T(x^\star) \leq c_1\sqrt{T} + c_2 \sqrt{\lambda T} e^{\lambda Q(T)/2} - \frac{1}{2}e^{\lambda Q(T)}.  
\end{eqnarray*}
Using the fact that $ax-bx^2 \leq \frac{a^2}{4b}, \forall b>0$, and taking $x=e^{\lambda Q(T)/2},$ the regret can be further upper bounded as follows:
\begin{eqnarray*}
	\textrm{Regret}_T(x^\star) \leq c_1 \sqrt{T} + \frac{c_2^2}{2} \lambda T = O(\max(\sqrt{T}, \lambda T)).
\end{eqnarray*}
Finally, choosing $\lambda = \min (\frac{1}{2c_3 B_T}, T^{-(1-\beta)})$ for some $0 \leq \beta \leq 1,$ we obtain the following trade-off
\begin{eqnarray} \label{reg-ccv-lt-bds}
	Q(T) = \tilde{O}(\max(B_T, T^{1-\beta})), ~\textrm{Regret}_T(x^\star) = O(\max(\sqrt{T}, T^\beta)).
\end{eqnarray}
As an example, if $B_T= O(T^{\nicefrac{1}{3}}),$ then by choosing $\beta = \nicefrac{2}{3},$ we obtain $\textrm{Regret}_T(x^\star) = O(T^{\nicefrac{2}{3}})$ and $\textrm{CCV}_T= \tilde{O}(T^{\nicefrac{1}{3}}).$  

The above results are summarized in the following Theorem. 
\begin{theorem} \label{cvx-smooth-lt-thm}
		Let $B_T \geq 0$ be the prescribed long-term constraint violation budget. Consider the Lyapunov function $\Phi(x)= e^{\lambda x}$ with $\lambda = \min (\frac{1}{c B_T}, T^{-(1-\beta)})$ where $\beta \in [0,1]$ is a tunable parameter and $c$ is a constant which depends on the problem parameters ($D$ and $M$) as discussed above. Then, under Assumptions \ref{bdd-assump}, \ref{cvx-smooth}, and \ref{feasibility-new}, Algorithm \ref{coco_alg3} achieves the following guarantee for any long-term feasible benchmark action $x^\star \in \mathcal{X}_{B_T}^\star$:
	\[\textrm{Regret}_T(x^\star) = O(\max(\sqrt{T}, T^\beta)), ~ \tilde{O}(\max(B_T, T^{1-\beta})).\]
\end{theorem}

\subsection{Experiments} \label{expt}
\begin{figure*}[h]
\centering
	\includegraphics[scale=0.35]{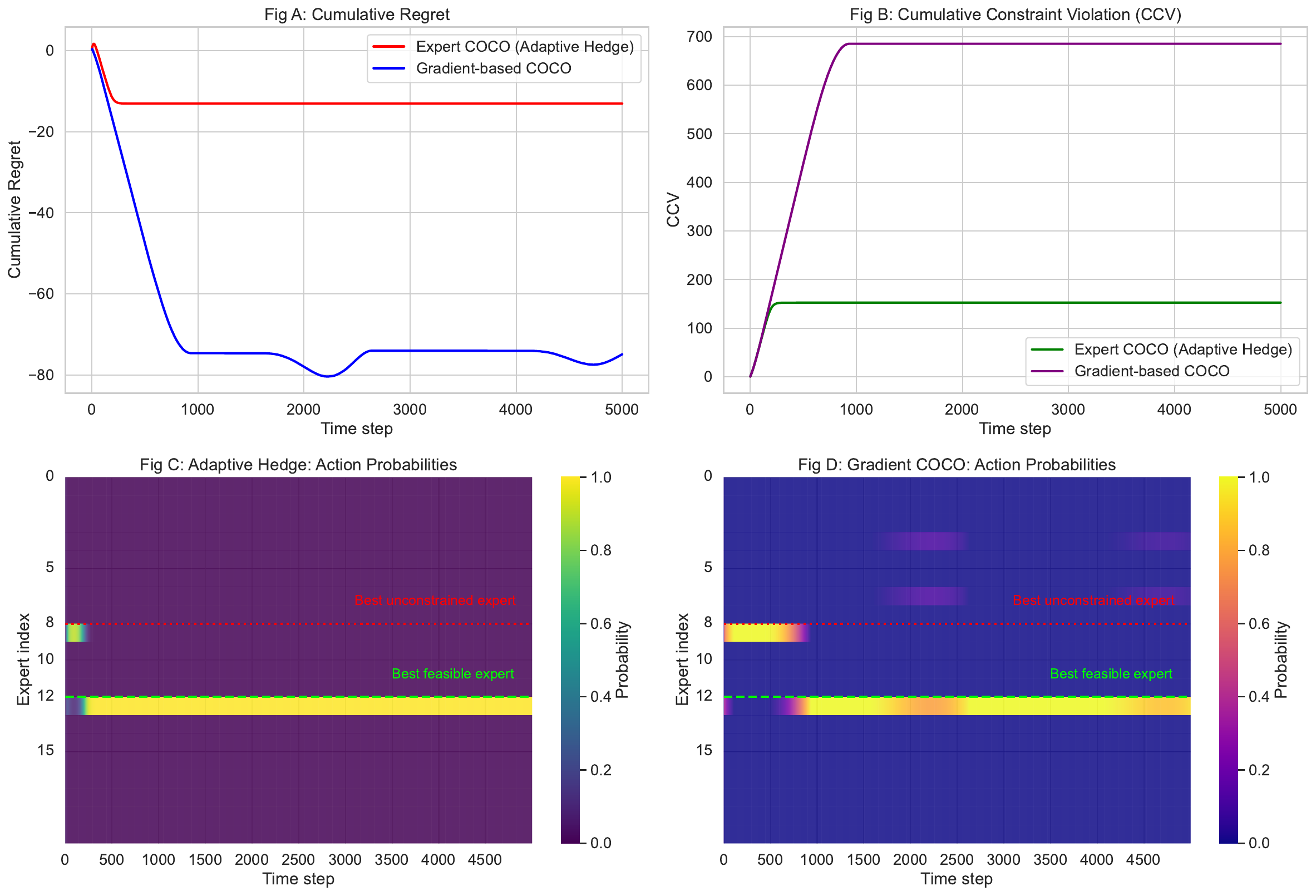}
	\caption{\small{Comparison of the proposed adaptive Hedge-based policy (Algorithm~\ref{coco_alg}) with $\beta = 0.75$ and the OGD-based policy from \citet[Algorithm 1]{sinha2024optimal}. 
    (A) Regret vs. time. 
    (B) Cumulative Constraint Violation (CCV) vs. time. 
    (C) Selection frequency of experts under our adaptive Hedge-based policy. 
    (D) Selection frequency of experts under the OGD-based policy.
    The proposed policy quickly identifies and sticks to the best feasible expert, leading to significantly lower CCV, whereas the OGD-based policy initially selects infeasible experts and takes longer to converge.}}
	\label{expt-plot}
\end{figure*}

\paragraph{Problem instance:} 
We consider the \textsc{Constrained Expert} problem on a synthetic dataset with \( N = 20 \) experts over a horizon of length \( T = 5000 \). Two experts are designated as special: the best feasible expert, denoted by \( \texttt{E}^\star \), and the best unconstrained expert, denoted by \( \texttt{UE}^\star \). The expert \( \texttt{E}^\star \) is feasible, with i.i.d.\ costs drawn from a distribution with mean \( \bar{f}_{\texttt{E}^\star} = 0.21 \), and zero constraint violation on all rounds, i.e., \( \bar{g}_{\texttt{E}^\star} = 0.0 \). In contrast, \( \texttt{UE}^\star \) is infeasible, with i.i.d.\ costs and constraint violations having a smaller mean \( \bar{f}_{\texttt{UE}^\star} = 0.11 \) and higher average constraint violation \( \bar{g}_{\texttt{UE}^\star} = 0.91 \), respectively. The remaining experts incur i.i.d.\ random costs with mean \( \bar{f} = 0.41 \) plus a zero-mean periodic component over time, and their constraint violations are i.i.d.\ with mean \( \bar{g} = 0.6 \). Additionally, two more experts, \( \texttt{DE}_1 \) and \( \texttt{DE}_2 \), distinct from both \( \texttt{E}^\star \) and \( \texttt{UE}^\star \), are made feasible by setting their constraint violations to zero on all rounds.
Table~\ref{param-table} summarizes the parameters used in the experiments. The experiments have been run on a quad-core CPU with 8 GB RAM. The source code has been made publicly available \citep{coco-code-improved}.

\begin{table*}[!ht]
  \title{Parameters for Experts Used in the Experiments}
  \centering
  \begin{tabular}{llll}
    \toprule
   \small {Expert(s)}  & \small{Index}& \small {Average cost} & \small {Average constraint violation}  \\
    \midrule
      $\texttt{E}^\star$ &$\# 12$ & $\bar{f}_{\texttt{E}^\star}=0.21$ & $\bar{g}_{\texttt{E}^\star}=0.0$ \\
          $\texttt{UE}^\star$ &$\# 8$ & $\bar{f}_{\texttt{E}^\star}=0.11$ & $\bar{g}_{\texttt{E}^\star}=0.91$ \\
          $\texttt{DE}_1, \texttt{DE}_2$ &$\# 3, \# 6$ & $\bar{f}_{\texttt{E}^\star}=0.41$ & $\bar{g}_{\texttt{E}^\star}=0.0$ \\
             \texttt{The rest} &$[20]\setminus \{3,6,8,12\}$ & $\bar{f}_{\texttt{E}^\star}=0.41$ & $\bar{g}_{\texttt{E}^\star}=0.6$ \\
       \bottomrule
  \end{tabular}
  \vspace{5pt}
  \caption{\small{Parameter settings for generating cost and constraint vectors for $N=20$ experts.} }
    \label{param-table}
\end{table*}

  \paragraph{Summary of the results.} 
Parts (A) and (B) of Figure~\ref{expt-plot} compare the performance of our proposed adaptive Hedge-based policy (Algorithm~\ref{coco_alg}) with $\beta=0.75$ with that of the Online Gradient Descent (OGD)-based policy proposed by \citet[Algorithm 1]{sinha2024optimal}. From the plots, it is evident that the proposed algorithm incurs significantly lower cumulative constraint violation (CCV) while maintaining a sub-linear regret. 

To gain deeper insight into the working of the algorithms, parts (C) and (D) of Figure \ref{expt-plot} show the relative frequency with which each expert is selected by our algorithm and the OGD-based policy, respectively. These plots clearly demonstrate that the adaptive Hedge-based policy quickly identifies the best feasible expert and predominantly selects it thereafter. In contrast, the OGD-based policy initially incurs a substantial amount of constraint violation by frequently selecting infeasible experts. Only after a considerable number of rounds does it converge to the best feasible expert and begin exploiting it consistently. 

\begin{figure}
\centering
	\includegraphics[scale=0.37]{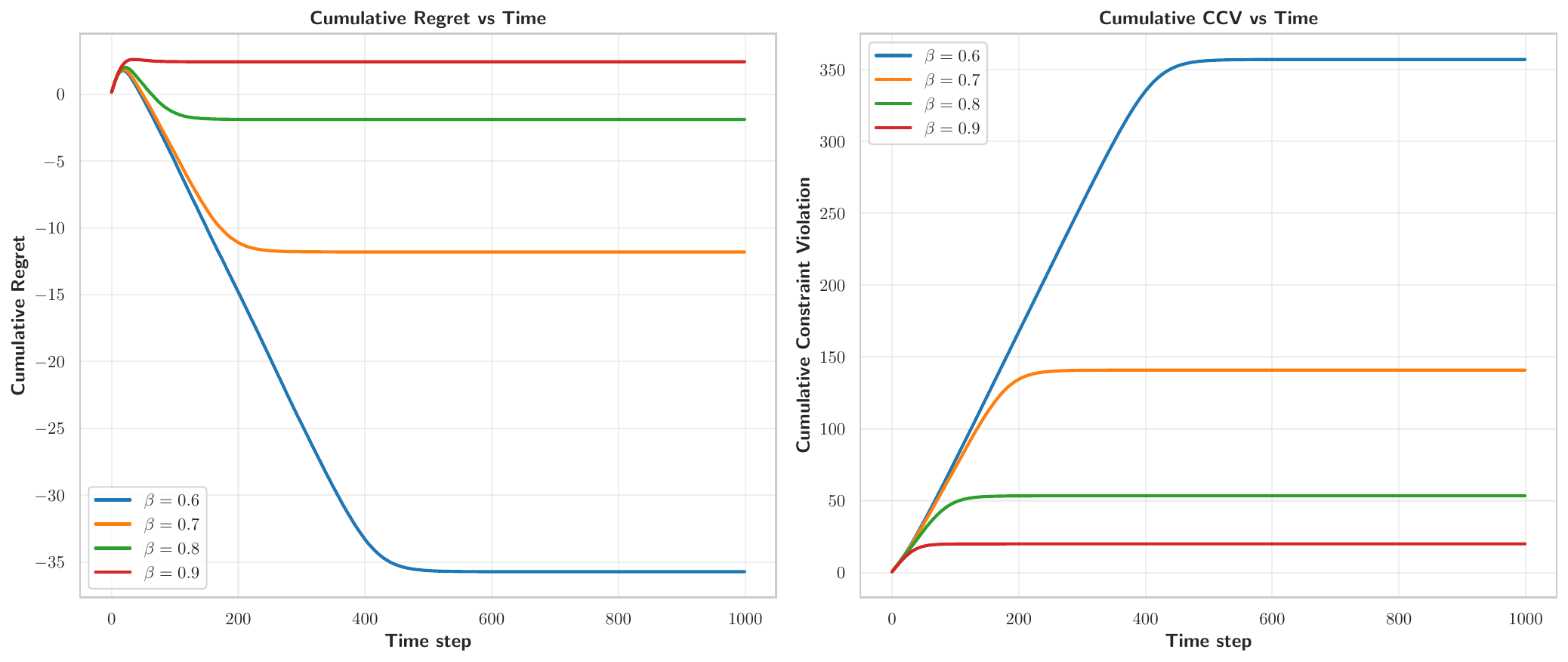}
	\caption{Performance comparison for different values of $\beta$}
	\label{beta-sweep}
\end{figure}

Figure~\ref{beta-sweep} illustrates the trade-off between regret and cumulative constraint violation (CCV) for different values of the tuning parameter \( \beta \) in the proposed policy. As \( \beta \) increases from 0.6 to 0.9, the algorithm incurs lower CCV at the expense of higher regret. This behavior aligns with the theoretical performance guarantees established in Theorem~\ref{main-th}.

\end{document}